\documentclass[11pt]{article} 

\usepackage[dvips]{graphicx}
\usepackage{amssymb}

\newtheorem{lemma}{Lemma}

\newtheorem{defi}{Definition}

\begin{document}

\begin{center}

{\Huge {\bf Gomoku: analysis of the game  \\ \vspace{6pt} and of the player Wine }}

\vspace{5pt} 

Lorenzo Piazzo, Michele Scarpiniti and Enzo Baccarelli \\
DIET dept. - Universita' di Roma 'Sapienza' \\
contact: \verb|lorenzo.piazzo@uniroma1.it|  \hspace{1cm}  \\ issue: october 31st, 2021 \\

\end{center}

\vspace{10pt}

{\bf Abstract. } Gomoku, also known as five in a row, is a classical board game, ideally suited for quickly testing novel Artificial Intelligence (AI) techniques. With the aim of facilitating a developer willing to write a new Gomoku player, in this report we present an analysis of the main game concepts and strategies, which is wider and deeper than existing ones. Moreover, after discussing the general structure of an artificial player, we present and analyse a strong Gomoku player, named Wine, the code of which is freely available on the Internet and which is an excelent example of how a modern player is organised.

\vspace{10pt}

\tableofcontents

\newpage
\section{Introduction}

Gomoku, also known as five in a row, is a simple, classical board game \cite{wiki,lasker}. It is known and played around the world but the largest players communities are in east Asia and east Europe. There exist several versions but the rules for the basic game are remarkably simple: a black and a white player alternately place a stone of their colour on a free square of a checkerboard; the first player able to place five stones in a row is the winner. Thanks to its simplicity the game can also be played using pencil and paper. 

While the rules are simple, the game is not: plenty of thinking and strategy is required in a match between two good players. This combination of simple rules and complex strategy makes Gomoku an ideal choice for the study of artificial intelligence (AI) techniques. In fact, the effort spent in programming the game rules is minimal, so that the developer can concentrate on the player's brain, which is the most interesting part. Indeed, developing an artificial Gomoku player is challenging, due to the high branching factor of the game, and only in recent times (2017) a computer program was able to win against a human professional player \cite{wiki}. As an additional benefit, a tournament for artificial players, called Gomocup \cite{gomocup}, is active and hosts plenty of good players. All these facts make Gomoku an excelent gym, where new ideas and AI concepts can easily and quickly be tested. 

As a drawback, the literature about the game strategies and the AI techniques used in the best players is scarce, at least in English. Concerning the game strategies, a seminal work is the one by Allis \cite{allis01}, where several important concepts are introduced. A few pages about Gomoku can also be found in the book by Lasker \cite{lasker}. Moreover, some basic strategic concepts can be found online, e.g. \cite{goworld}. Concerning the players, neither the code nor the design choices of the top ones have been disclosed. However, the code of some good players has been made publicly available, e.g. \cite{carbon,wine,alphagomoku}, but without any documentation, except for the comments in the code. As a result, a developer willing to write a Gomoku player has little material to start with. His best option is to perform a reverse engineering study of the code of one of the disclosed players, which is not a simple task.

To overcame the above mentioned drawbacks, in this report we present an analysis of the game that is wider and deeper than the exisiting ones. The analysis is considered a work in progress, because we feel that it can be improved. However, it surely introduces concepts that are needed in any development and considers several aspects that are useful when developing an artificial player. In addition, we provide a discussion of Wine, which is a player developed by Jinjie Wang, the code of which is available on Github \cite{wine}. Wine is a strong player: it ranked tenth in the 2020 edition of the free style tournament of the Gomocup. Moreover, the code is clear and well written and the brain is comparatively simple. As a final contribution, we report some understandings gained in the last year, which we spent developing a new player that will hopefully compete in the Gomocup soon or later. These include some improvements that can be made to the Wine code.

The report is orgainsed as follows. In section \ref{sec.basic}, we carry out an analysis of Gomoku, by introducing several basic concepts and results. In section  \ref{sec.player} we discuss the general structure of an artificial player and review several different design options. In section \ref{sec.wine} we describe the Wine player. 

{\bf Notation.} In the report we need many definitions. The most important are explicitely noted and numbered, but, in order to simplify the reading, many are implicit: in the latter case the terms being defined are written in boldface.

\newpage
\section{Gomoku}
\label{sec.basic}

Gomoku has several variants. Here we only consider the simplest one, which is called free style Gomoku. It is played on a checkerboard of $N_r$ rows by $N_c$ columns by two players\footnote{In practice, the game is played at the intersections of a grid of $N_r$ by $N_c$ lines, but using a checkerboard is entirelly equivalent.}. Tipically the board is squared and $N_r=N_c=15$ or $19$. The board's squares are indexed, from $1$ to $N_r N_c$. The two players are black (PB) and white (PW). The black player places the first (black) stone on the board. After that, the two players alternately place a stone of their colour in a free square of the board. The first player making a line (horizontal, vertical or diagonal) of five consecutive stones of the same colour is the winner. If the board is filled up without making a five, the game is a draw.

We note that in \cite{allis01} Allis showed that the first player can always win the free style game on a $15$ by $15$ board. A similar result is expected to hold on wider boards too. However, this is a theoretical result. Indeed, in a real match, a player cannot follow the winning strategy due to time and memory constraints. In practice, it routinely happens that the second player wins the game, even in a match between top players (artificial or human). However, the first player does have an advantage which translates into a higher winning probability. To counter this problem, several ways exist. The simplest one is to consider a match as two games with swapped side. Other ways include putting restrictions on the first player, like prohibiting to place its second stone near the first one, or prohibiting to place its first stone near the board center. We will not consider such variations in this report, but the analysis that we present is largely applicable to these cases too. 

In the next subsection, we introduce several basic concepts and definitions that will be useful in the game analysis.

\subsection{Players and patterns}

\begin{defi} \label{def.rational} {\em {\bf Rational player:}  a player that, if can win, wins in the quickest possible way (least moves) and, if has to loose, looses in the slowest possible way (most moves). } $\blacklozenge$ \end{defi}

From now on, we assume that both players are rational players. We refer to a generic player as to PX. The other player is PY. The board status before the PX's move will be termed a {\bf PX board}. In other words, on a PX board, the next stone placed is a PX stone.

\begin{defi} \label{def.pattern} {\em {\bf Pattern:} a subset of the board's squares together with the specification of the contents of each square (black, white, empty). The number of squares in the pattern is said the {\bf size} of the pattern.} $\blacklozenge$ \end{defi}

We will use greek letters to indicate patterns. A pattern $\pi$ can be depicted as illustrated below (where the \verb|X| is a stone of PX, \verb|O| is a stone of PY and \verb|+| is an empty square):

\linespread{0.6}   
\begin{verbatim}
X++O++O+          a pattern on two rows
   O++X++
\end{verbatim}
\linespread{1.1}   

Note that in the picture above, we are not explicitely giving the board's squares hosting the pattern. We are merely giving the shape and contents of the pattern. Such a representation can be called a relative pattern. Most of our reasonigs only require relative patterns. If an absolute pattern has to be specified, we will write the index number of one of the squares near it, in order to pin the pattern on the board. For example:

\linespread{0.6}   
\begin{verbatim}
i
X++O++O+          an absolute pattern, with leftmost square number i
   O++X++
\end{verbatim}
\linespread{1.1}   

We review some basic and obvious operations that can be applied to a pattern. Given a pattern $\pi$, by selecting a subset of its squares we obtain a second pattern $\sigma$ which is said a {\bf subset} of $\pi$, written $\sigma \subset \pi$. We also say that $\pi$ {\bf contains} $\sigma$. In particular, we will often perform the operation of {\bf removing} a square from a pattern of size $n$, which produces pattern of size $n-1$. Another useful operation that we will use is the {\bf empting} of one of the filled squares of the pattern or the {\bf filling} of one of the empty squares. These operations do not change the pattern size, only its contents.

Given two patterns, we say that the two are {\bf compatible} if the contents of the common squares are identical. This means that the patterns can be on the board at the same time. Given two compatible patterns we define their {\bf union and intersection} in the obvious way, written, respectively, $\alpha \cup \beta$ and $\alpha \cap \beta$. As a final definition, when a pattern contains no PY stones, we say that it is a {\bf PX Pattern}.

\vspace{6pt} \noindent
{\bf Restricted game.} Given a pattern $\pi$ we can consider a game that is played in the pattern's squares only. Such a game is termed the game restricted to the pattern. Beside being restricted, the game follows the same rules of the full game. In particular, there is a first player, which can be either PX or PY and needs to be specified. Starting with the first, the players alternately place a stone in one of the empty squares. The winner is the first player making a line of five and if the pattern gets filled with no winner the game is a draw.

\vspace{6pt} \noindent
{\bf Block.} A pattern constituted by five aligned and consecutive squares (horizontally, vertically or diagonally) will be called a block. If the block contains stones of both PB and PW it is said a {\bf dead } block. If the block is not dead, it is said a {\bf PX block} if it contains stones of PX and its {\bf degree} is the number of stones that it contains. A PX block of degree $n$ will be indicated by BX$n$ or simply by B$n$ when the owner is clear.

\vspace{6pt} 
In the next subection, we introduce some important patterns and concepts, that will be used throughout the report.

\subsection{Victories, attacks and threats}

\begin{defi} \label{def.victory} {\em {\bf PX Victory:} a PX pattern such that PX wins the restricted game if he is either the first or the second player of the restricted game. } $\blacklozenge$ \end{defi}

A victory is clearly an important pattern. In fact, if a player is able to place a victory on the board, he can win the full game. This will be better discussed in section \ref{vic.ana}.

Given a PX victory, the number of stones that PX has to place in order to achieve the victory when he is the second player of the restricted game is called the number of {\bf steps} of the victory. A victory of $n$ steps is denoted by V$n$. Sometimes we also want to specify the owner: in this case we write VX$n$ and VY$n$. Here are some victory examples:

\begin{verbatim}
XXXXX       PX victory of 0 steps (VX0) 
+XXXX+      PX victory of 1 step (VX1)
+OOOO+      PY victory of 1 step (VY1)
+XXXX++     PX victory of 1 step (VX1)
XXXX++      not a victory (PX wins only if he plays first)
\end{verbatim}

\linespread{0.6}   
\begin{verbatim}
XXXX+       PX victory of 1 step (VX1)
X              
X
X
+
\end{verbatim}
\linespread{1.1}

It is convenient to give names and symbols to some patterns. We will do this in a structured way in section \ref{pat.exe}, but we start here using the comments near the figures. Here are the first two:

\begin{verbatim}
XXXXX       simple five (S5) - a V0
+XXXX+      double four (D4) - a V1
\end{verbatim}

\begin{defi} \label{def.attack} {\em {\bf PX Attack:} a PX pattern such that 1) it is not a victory and 2) it contains one or more empty squares, termed the {\bf triggers}, such that when one of these squares is filled with a PX stone the pattern becomes a PX victory. } $\blacklozenge$ \end{defi}

An attack is a second important pattern. In fact, since it can be turned into a victory, PX wins the restricted game if he is the first player, i.e. when the attack is on a PX board. On the other hand, since the pattern is not a victory, PX cannot win the restricted game if he is the second player. i.e. if the attack is on a PY board. However, in the latter case, PY shall stop the attack and has therefore a limited set of moves, which is an important advantage.

Consider a PX attack and its triggers. The number of {\bf steps} of the attack is one plus the number of steps of fastest (least steps) victory that can be produced by playing in one of the triggers. It follows that the number of steps is greater than or equal to one. An attack of $n$ steps is denoted by A$n$. Sometimes we also want to specify the owner: in this case we write AX$n$ and AY$n$. Here are some examples:

\begin{verbatim}
XXXX+       PX attack of 1 step - A1 - simple four (S4)
XX+XX       PX attack of 1 step - A1 - simple four (S4)
+XXXX+      Not an attack (it is a victory)
\end{verbatim}

When there are several empty squares we may want to indicate the triggers. We do this by placing a \verb|*| near the trigger square. Note that an attack may contain more than one trigger. Some examples follow:

\linespread{0.6}   
\begin{verbatim}
++XXX+      PX attack of 2 steps - A2 - weak three (W3)
 *

++XXX++     PX attack of 2 steps - A2 - double three (D3)
 *   *
\end{verbatim}
\linespread{1.1}

\begin{defi} \label{def.threat} {\em {\bf PX Threat:} a PX pattern such that 1) it is not a victory nor an attack and 2) it contains one or more empty squares, termed the {\bf triggers}, such that when one of these squares is filled with a PX stone the pattern becomes a PX attack.} $\blacklozenge$ \end{defi}

A threat is an additional useful pattern. In fact, when a PX threat is on a PX board, by playing in its trigger PX will place an attack on the PY board, thereby limiting the moves of PY.

Consider a PX threat and its triggers. The number of {\bf steps} of the threat is one plus the number of steps of the fastest (least steps) attack that can be obtained by playing one of the triggers. It follows that the number of steps is greater than or equal to two. A threat of $n$ steps is denoted by T$n$. Sometimes we also want to specify the owner: in this case we write TX$n$ and TY$n$. Here are some examples:

\linespread{0.6}   
\begin{verbatim}
+XXX+      PX threat 2 steps, 2 triggers - T2 - simple three (S3)
*   *

++X+X++    PX threat 3 steps, 3 triggers - T3 
 * * *
   
+++XX++++++XX+++   T3 with many triggers (not all indicated)
  *  *    *  *

XXX+++++++XX+++   T2 with many triggers (not all indicated)
   *     *  *

XXXX++++++XX+++   Not a threat (it is an A1)
\end{verbatim}
\linespread{1.1}

\vspace{6pt} \noindent
{\bf Generic pattern.} We have introduced three types of patterns, namely victories, attacks and threats. Note that, based on the definitions, the three classes are disjoint, that is, a pattern cannot be, for example, a victory and an attack at the same time. Moreover, any pattern which does not belong to one of these classes will be called a {\bf generic} pattern.

\vspace{6pt} \noindent
{\bf Defence of an attack.} Consider a PX attack $\alpha$. In the pattern it must exist at least one empty square such that if this square is removed the pattern becomes a threat or a generic pattern\footnote{In fact, suppose that there is no such square and that PY plays first in the restricted game. We can remove the square played by PY and still have an attack. This means that PX can win even if he plays second, i.e. that $\alpha$ is a PX victory: but this is impossible by definition \ref{def.attack}. }. Such squares will be called the {\bf defence} of the attack. Indeed, if PY is the first player of the restricted game, he can stop the attack by playing in the defence. The number of these squares is the {\bf size} of the defence. Below, we present some examples where we mark a defence square using a dash \verb|-| near it:

\linespread{0.6}   
\begin{verbatim}
    -
XXXX+       the simple four has a size 1 defence
    *

 -   -
++XXX++     the double three has a size 2 defence
 *   *

--   -
++XXX+      the weak three has a size 3 defence
 *
\end{verbatim}
\linespread{1.1}   
 
\vspace{6pt} \noindent
{\bf Defence of a victory and a threat.} The defence can be introduced for victories and threats too. In particular, a victory can be considered to have  a {\bf zero size defence}. Concerning a threat $\tau$, recall that it has a set of triggers and that, for any trigger $\gamma$, PX can make an attack $\alpha$ by playing in $\gamma$. The defence of $\alpha$ will be said the {\bf defence of $\tau$ for the trigger $\gamma$}. Thus, in a threat, there is not a single defence, but a defence for each trigger. However, it is convenient to define a defence {\em size} applicable to the whole threat too. To this end, recall that the number of steps of the threat is dictated by the fastest of the attacks that can be triggered. Among the fastest attacks, there will be one or more with the smallest defence size. That size is considered the {\bf defence size of the threat $\tau$}. 

\linespread{0.6}   
\begin{verbatim}
    -
+XXX+      a T2 - a trigger and its defence marked
*   

-
+XXX+      same T2 - other trigger and its defence marked
    *   

 -   -
++X+X++    a T3 - a trigger and its defence marked  
   *

 - -  - 
++X+X++    same T3 - another trigger and its defence marked  
     *

\end{verbatim}
\linespread{1.1}   

\vspace{6pt} \noindent
{\bf Strength of a pattern.} It is convenient to assign a strength to each pattern. To this end, we look to the type, number of steps and size of the defence, in that order. In particular, a victory is stronger than an attack, an attack is stronger than a threat and a threat is stronger than a generic pattern. For patterns of the same type, the one with the least number of steps is the stronger. And when type and number of steps are the same, the one with the smallest defence is the stronger. The generic patterns all have the same strength.

\vspace{6pt} \noindent
{\bf Minimal, impossible and irrational patterns.} The game analysis can be simplified by ignoring some trivial patterns. To this end, in the following we introduce the concepts of minimal, impossible and irrational patterns. 

\vspace{6pt} \noindent
{\bf Minimal pattern.} Consider a non generic PX pattern $\pi$. By removing a square from it, we obtain a second pattern $\beta$. If, for any square removed from $\pi$, the resulting pattern $\beta$ is weaker than $\pi$, then $\pi$ is said a minimal pattern. Below, some examples.

\begin{verbatim}
XXXXX     minimal V0       +XXXX+    minimal V1        XXXX+    minimal A1 
XXXXX++   non minimal V0   +XXXX+X   non minimal V1    XXXX++   non minimal A1
\end{verbatim}

In the analysis, we can ignore non minimal patterns and concentrate on the minimal ones. In fact, any non minimal pattern of a given strength will contain a simpler, minimal pattern of the same strength. 

\vspace{6pt} \noindent
{\bf Impossible pattern.} There are some patterns that cannot appear in a game. For example, any pattern containing two disjoint S5, e.g. \verb|XXXXX+XXXXX|, cannot appear, because the game would end when the first S5 is placed on the board. These patterns can be ignored in the analysis. 

\vspace{6pt} \noindent
{\bf Irrational pattern.} There are some patterns that are impossible or unlikely when the players are rational. These patterns will be said irrational and can be ignored in the analysis. In order to show that a PX pattern is irrational, we must show that PX has a better strategy than making that pattern. For example, consider the following pattern: \verb|XXXX+XXXX|. While the pattern, strictly speaking, is possible, no rational player would ever produce it. To see why, just remove any PX stone and note that PX could win the game by placing that stone in the central square. The latter example is a special case of a class of irrational patterns which is identified in the following definition.

\begin{defi} \label{lemma.irra} {\em {\bf Irrational Double Attack:} a PX pattern containing two or more attacks such that the attacks' intersection contains no filled squares. } $\blacklozenge$ \end{defi}

In order to understand the rationale of the definition, note that since the two attacks do not intersect in any filled square, they do not have any common stone. Therefore, when the second is placed on the board, the first one was already there. However, a rational player would not place the second attack: he would play in the trigger of the first attack and achieve a victory. Therefore such patterns are impossible or unlikely\footnote{There are situtations in which PX has an attack on the board but cannot play it and is forced to place a second, disjoint attack on the board. This may happen, for example, when PX has to stop a faster PY attack.}. 

Having introduced the main definitions, in the next subsection we present several Lemmas that will be useful in the analysis.

\subsection{Useful Lemmas}
\label{sec.constr}

The following two Lemmas state that we can construct an attack from a victory and a threat from an attack. The proofs are obvious and stem directly from the definitions.

\begin{lemma} \label{lemma.attack} {\em {\bf Simple Attack.} Consider a PX victory of $n$ steps. By empting any square containing a stone, we obtain a pattern which is a PX attack of $n+1$ steps or stronger. } $\blacklozenge$ \end{lemma}

\begin{lemma} \label{lemma.threat} {\em {\bf Simple Threat.} Consider a PX attack of $n$ steps. By empting any square containing a stone, we obtain pattern which is a PX threat of $n+1$ steps or stronger.} $\blacklozenge$ \end{lemma}

For example, starting from a simple five \verb|XXXXX|, which is a V0, we can produce five simple fours, which are A1, namely \verb|+XXXX|, \verb|X+XXX|, \verb|XX+XX|, \verb|XXX+X| and  \verb|XXXX+|. And starting from a simple four, e.g. \verb|XXXX+|, which is an A1, we can produce four simple threes, which are T2, namely \verb|+XXX+|, \verb|X+XX+|, \verb|XX+X+|, \verb|XXX++|. 

Note that, in some cases, we obtain a stronger pattern than the one predicted by the Lemmas, but such cases are trivial. For example, consider the following non minimal V1: \verb|+XXXX+X|. By empting the last square we obtain a V1 and not an A2. However, the new V1 is of no interest, because it is essentially the same V1 of the original pattern. Most of these trivial cases can be ruled out by restricting to minimal and rational patterns. 

\vspace{6pt}

The following Lemma state that we can construct a victory from two attacks having disjoint defences.

\begin{lemma} \label{lemma.vic.comb} {\em {\bf Combination Victory.} Consider two compatible PX attacks $\alpha_1$ and $\alpha_2$ with steps $n_1$ and $n_2$ respectively. If the defences are disjoint, the union of $\alpha_1$ and $\alpha_2$ contains a victory. The number of steps of the victory is at most $n = max( n_1, n_2)$.

\noindent
{\bf Proof:} Consider the union $\mu = \alpha_1 \cup \alpha_2$. It is a PX pattern containing $\alpha_1$ and $\alpha_2$. If PX plays first, he can play the trigger of an attack and win. If PY plays first, he can play in the defence in order to stop the attacks. However, since the defences are disjoint, PY cannot stop both attacks. A rational player will stop the quickest one. Then PX triggers the slower attack and wins. Therefore $\mu $ contains a PX victory, having the same number of steps of the slowest attack. } $\blacklozenge$ \end{lemma}

As a comment, note that {\em the two attacks must intersect in one or more filled squares}, otherwise the pattern would be a trivial, irrational double attack. Let us examine an example. Consider the following two A1: \verb|XXXX+| and \verb|+XXXX|. Suppose that their intersction is constituted by the four filled squares. Then, the two are compatible and the defences disjoint. Their union is \verb|+XXXX+| which is indeed a V1.

\vspace{6pt}

As we have seen, the union of two attacks with disjoint defences produces a victory. If the defences are not disjoint, the union of two attacks can anyway produce a better attack, with a smaller defence. This is stated in the following Lemma.

\begin{lemma} \label{lemma.att.joint} {\em {\bf Intersection Attack.} Consider two compatible PX attacks $\alpha_1$ and $\alpha_2$ with steps $n_1$ and $n_2$ and defences $\delta_1$ and $\delta_2$, respectively. Suppose that the intersection of the defences $\delta = \delta_1 \cap \delta_2$ is not empty. Then, the union $\mu = \alpha_1 \cup \alpha_2$ contains an attack with defence $\delta$ and with a number of steps $n = min ( n_1, n_2)$.  

\noindent
{\bf Proof:} Consider the union $\mu$. It is a PX pattern containing $\alpha_1$ and $\alpha_2$. If PX plays first, he can play the trigger of the quickest attack and win. If PY plays first, he can stop both attacks, provided that he plays in $\delta$. Therefore $\mu$ contains an attack with steps $n = min ( n_1, n_2)$ and defence $\delta$. } $\blacklozenge$ \end{lemma}

As an example, below, we show two compatible\footnote{We give the position of square $i$, so that the intersection can be computed and compatibility checked.} weak three. Each of the W3 has a size three defence; their union is a double three, having a size two defence.

\linespread{0.6}   
\begin{verbatim}
  W3             W3             D3
-   --         --   -          -   -
+XXX++         ++XXX+         ++XXX++
  i *           * i            * i *
\end{verbatim}
\linespread{1.1}

\vspace{6pt}

The following Lemma state that we can construct an attack from two threats having a common trigger.

\begin{lemma} \label{lemma.att.comb} {\em {\bf Combination Attack.} Consider two compatible PX threats $\tau_1$ and $\tau_2$, having a common trigger $\gamma$. Denote by $\delta_1$ and $\delta_2$ the defences associated with the trigger $\gamma$ in $\tau_1$ and $\tau_2$,  respectively. Denote by $n_1$ and $n_2$ the number of steps of the attacks triggered by $\gamma$ in $\tau_1$ and $\tau_2$,  respectively. If the two defences are disjoint, the union $\mu = \tau_1 \cup \tau_2$ contains an attack. One of the triggers of the attack is $\gamma$. The defence of the attack is $ \delta = \gamma \cup \delta_1 \cup \delta_2$. The number of steps of the attack is $n = 1 + max(n_1, n_2)$.

\noindent
{\bf Proof:} Assume that PX plays first. By playing in $\gamma$, PX will place on the board two attacks with disjoint defences. From Lemma \ref{lemma.vic.comb} this is a victory and recalling definition \ref{def.attack} we conclude that $\mu$ contains an attack. Since PY is a rational player he will stop the quickest attack, by playing into its defence. Therefore the attack will take $n = 1 + max(n_1, n_2)$ steps. Now assume that PY is the first to play. It he plays in $\gamma$ or $\delta_1$ or $\delta_2$, PX will not be able anymore to play the trigger and make a victory. Therefore the defence of the attack is $ \delta = \gamma \cup \delta_1 \cup \delta_2$. } $\blacklozenge$ \end{lemma}

As an example, below we consider two T2 with a common trigger (in square number $i$) and disjoint defences, and contruct an A2.

\linespread{0.6}   
\begin{verbatim}
  T2            T2             A2
   -i         i-               ---
XXX++         ++XXX         XXX+++XXX
    *         *                 *
\end{verbatim}
\linespread{1.1} 

\vspace{6pt}

Finally, the following Lemma is the inverse of the Combination Victory Lemma.

\begin{lemma} \label{lemma.dec} {\em {\bf Victory Decomposition.} A PX victory $\pi$ with a number of steps $n \geq 1$ contains at least two attacks with disjoint defences and a number of steps lower than or equal to $n$.

\noindent
{\bf Proof:} Suppose that $\pi$ contains no attacks. Then PX cannot win if he is the first player, but this is impossible because $\pi$ is a victory. Suppose that $\pi$ contains one attack only. Then PX cannot win if he is the second player, because PY would play in the attack defence, but this is impossible because $\pi$ is a victory. Therefore $\pi$ must contain at least two attacks and their defences must be disjoint, otherwise PY could stop both attacks when he is the first player. } $\blacklozenge$ \end{lemma}

As an example, below, we decompose a double four (a V1) into two simple fours (A1s) with disjoint defences. We use the \verb|=| symbol to mark the constituting attacks.

\linespread{0.6}   
\begin{verbatim}

 =====      S4       
 +XXXX+     D4       The D4 contains two S4
  =====     S4
\end{verbatim}
\linespread{1.1}




\subsection{Basic patterns}
\label{pat.exe}

In this section we give some examples and review several important patterns of one and two steps. In the following it is understood that we speak of patterns belonging to the PX player.

\noindent
{\bf $\bullet$ Simple five, S5.} The simple five (S5) is a pattern constituted by a single B5, a block of degree 5. Therefore the pattern is made by five aligned and consecutive stones: \verb|XXXXX|. It is a V0. It is the only minimal V0.

\noindent
{\bf $\bullet$ Simple four, S4.} The simple four (S4) is a pattern constituted by a single B4, a block of degree 4. Therefore the pattern is made by five aligned and consecutive squares, four filled and one empty, e.g. \verb|XXXX+|. The pattern can be obtained from an S5 by using the Simple Attack Lemma. It is an A1. It is the only minimal A1. The empty square is both the trigger and the defence. 

\noindent
{\bf $\bullet$ Simple three, S3.} The simple three (S3) is a pattern constituted by a single B3, a block of degree three. Therefore the pattern is made by five aligned and consecutive squares, three filled and two empty, e.g. \verb|+XXX+|.  The pattern can be obtained from an S4 by using the Simple Threat Lemma. It is a T2 with two triggers. It is a minimal pattern.

\noindent
{\bf $\bullet$ Double four, D4.} The double four (D4) is a pattern such that: it is minimal; it contains no S5; it contains two S4 with disjoint empty cells. Since it contains two attacks with disjoint defences, from the Combination Victory Lemma, it is a V1. It is the only minimal V1. Below some examples:

\linespread{0.6}   
\begin{verbatim}
+XXXX+            A D4 - contains two S4

+XXXX++           Not a D4 (not minimal) but contains a D4

XXX+X+XXX         A D4 - contains two S4 and several S3      

XXX+X             A cross D4 - Also denoted as C44
    + 
    X
    X
    X

XXX+X+XXX         Not a D4 (not minimal) but contains three D4
    + 
    X
    X
    X
  
\end{verbatim}
\linespread{1.1}

\noindent
{\bf $\bullet$ Weak three, W3.} The weak three (W3) is a pattern such that: it is minimal; it contains no S5 nor S4; it contains two S3 such that they intersect in one and only one empty square (and in any number of filled squares). Since the S3 is a T2, from the Combination Attack Lemma we see that the W3 is an A2, with the trigger in the empty square where the two S3 intersects and with a defence constituted by three empty squares (the intersaction and the other two empty squares). Moreover, the W3 can be obtained from a D4 by using the Simple Attack Lemma. Below, some examples:

\linespread{0.6}   
\begin{verbatim}
+X+XX+            A W3 - contains two S3

+X+XX++           Not a W3 (not a minimal pattern) but contains a W3

XXX+++XXX         A W3 - contains two S3 and several B2      

XXX++             A cross W3  
    + 
    X
    X
    X
\end{verbatim}
\linespread{1.1}

\noindent
{\bf $\bullet$ Double three, D3.} The double three (D3) is a pattern such that: it is minimal; it contains no S5 nor S4; it contains two W3 such that they intersect in two and only two empty squares (and in any number of filled squares). It is an A2 having two triggers (the same of the W3s). By using the Intersection Attack Lemma we see that the defence is made by the two intersecting empty squares. Below, an example:

\linespread{0.6}   
\begin{verbatim}
======              W3
 *   *
++XXX++             D3
 -   -   
 ======             W3
\end{verbatim}
\linespread{1.1}

\noindent
{\bf $\bullet$ Some irrational patterns.} In the leftmost part of the plot below, we present another D3, but we note that this pattern is an irrational double attack, since it contains two W3 intersecting in empty cells only. So this case is of little interest. Another irrational pattern is one made by two W3 intersecting in a single empty cell, like the one depicted on the rightmost part of the plot. Using the Intersection Attack Lemma, it is an A2 with a single defence square. However, since it is an irrational pattern, it is not worth a name.

\linespread{0.6}   
\begin{verbatim}
======              W3                          ======               W3
    **                                              ***  
+XXX++XXX+          D3                          +XXX+++XXX+                                       
    --                                               -        
    ======          W3                               ======          W3
\end{verbatim}
\linespread{1.1}

\noindent
{\bf $\bullet$ Cross four C44.} A special case of D4 is obtained by intersecting two S4 on different lines. The corresponding pattern will be called a cross four (C44). Below, two examples:

\linespread{0.6}   
\begin{verbatim}
   C44                                 C44
   
XXX+X                                   X
    +                                   +
    X                                 XXX+X
    X                                   X
    X                                   X
\end{verbatim}
\linespread{1.1}

\noindent
{\bf $\bullet$ Cross four three C43.} By using the Combination Victory Lemma we can combine an S4 (which is an A1) with a D3 or a W3 (which are A2) on two different lines. If the attacks have disjoint defences we obtain a V2 that will be called a cross four three (C43). When needed, we can track if the three is weak or double and write C43w or C43d. Below, two examples:

\linespread{0.6}   
\begin{verbatim}

C43w                      C43d

    +                        +
XXX+X                        +
    +                    XXX+X
    X                        X
    X                        X
    +                        +
                             +
\end{verbatim}
\linespread{1.1}

\noindent
{\bf $\bullet$ Cross three C33.} By using the Combination Victory Lemma  we can combine two D3 or W3 (which are A2) on different lines. If the attacks have disjoint defences we obtain a V2, that will be called a cross three (C33). When needed, we can track if the three are weak or double and write C3w3w, C3w3d etc. Below, two examples:

\linespread{0.6}   
\begin{verbatim}

C3w3w                    C3w3d

    +                        +
+XX+X+                       +
    +                    +XX+X+
    X                        X
    X                        X
    +                        +
                             +
\end{verbatim}
\linespread{1.1} 

\noindent
{\bf Comment.} It is worth noting that, in a match between two rational players, the winning player has to place a D4 on the PY board in his second last move. In fact, if he placed an S4 only, PY would play in the S4 defence and negate the win. In turn, this implies that the winning player has to play in the trigger of a W3 in his second last move.

\subsection{Complex patterns}
\label{pat1.exe}

In the last section, we reviewed all the interesting one and two steps patterns. Clearly, there are attacks and victories with more than two steps. However, these are many more and a full inventory makes little sense. Never the less, we want to show some examples and discuss how these complex patterns can be constructed, because these techniques could be useful when developing an artificial player. In particular, in this section we will present two construction methods. The first one is based on the Lemmas given in section \ref{sec.constr}. The second one is based on the concept of threat sequence.

\vspace{6pt}
\noindent
{\bf Construction based on the Lemmas.} We proceed to show how the Lemmas presented in section \ref{sec.constr} can be used to produce new, complex patterns. At the same time, we will discuss some aspects of the Lemmas. We start by using the Simple Attack Lemma, which states that we can produce an attack of $n+1$ steps or a stronger pattern from a victory of $n$ steps by empting a square. Consider for example the C3w3w that was presented earlier, which is a V2. We want to use the Lemma to produce a novel A3. Below we report the V2 (on the left) and two attacks that can be produced by empting one of its squares.

\linespread{0.6}   
\begin{verbatim}
   V2         A3 (trigger in the intersection)     A2  
   +                        +                     +
+XXX++                   +XX+++                +XXX++       
   +                        +                     +
   X                        X                     +  
   X                        X                     X  
   +                        +                     +                     
\end{verbatim}
\linespread{1.1} 

The first attack, in the middle, is an A3 and it is a novel, minimal pattern. The second one on the right, is an A2. Thus, in the latter case, the Lemma produced an attack which is even stronger than an A3 but is trivial, since it is just non mimimal W3. 

The latter example can be generalised. Consider a victory $\pi$. From the Victory Decomposition Lemma, $\pi$ contains at least two attacks. Moreover, if $\pi$ is rational, the attacks must intersect in one or more non empty squares. If we want to produce a non trivial, novel attack from $\pi$ using the Simple Attack Lemma, we must empty one of these intersection squares, otherwise we would obtain just a non minimal version of one of the two contained attacks.

Since we produced a new A3, we can use the Combination Victory Lemma to produce a new V3. For example, we can combine the new A3 with a W3, being careful to intresect the two in a non empty square, in order to avoid irrational patters, to obtain the V3 shown on the left in the plot below. Next, we can use the Simple Attack Lemma again and empty the intersection square to produce a novel attack, which is an A4, shown on the right plot below.

\linespread{0.6}   
\begin{verbatim}
  V3                 A4 (trigger in the left intersection)
 +  +                        +  +
+XX+++                      ++X+++ 
 X  +                        X  +
 X  X                        X  X
 +  X                        +  X
 +  +                        +  +         
\end{verbatim}
\linespread{1.1} 

In summary, by repeatedly applying the Combination Victory and the Simple Attack Lemmas, we can produce victories and attacks with more and more steps. Moreover, for each of the attacks, we can use the Simple Threat Lemma and obtain a corresponding threat. And by using the Combination Attack Lemma we could use these threats to produce additional attacks.

\vspace{6pt}
\noindent
{\bf Construction based on threat sequences}. A classical Gomoku tactic is based on a complex pattern known as a threat sequence \cite{allis01,goworld}. In this pattern PX plays a threat every move, thereby placing an attack on the PY board. In this way, PY has a limited set of moves and PX may stay in control of the game. If the threat sequence ends up in a victory, PX can win the game. Below we report a simple example involving a sequence of S3. In the example, PX can win the restricted game on the left of the plot if he is the first player, by playing in the squares number 1 and 2. PY has no choice at all. Indeed, in every PY board an S4X is there and PY must play in its defence to avoid an immediate defeat. After two PX and two PY moves, the PX board on the right is obtained, containing a D3X, and PX wins with two more moves. Therefore the pattern on the left is an A4.

\linespread{0.6}   
\begin{verbatim}
    1
XXX++                           XXXOX                     
    X                               X                           
    X                               X                        
    +                               O                        
 ++X+X++                         ++XXX++                                       
    2
\end{verbatim}
\linespread{1.1}

A second example is reported below. The example is similar to the previous one. The only difference is that, in his first move, PX plays a T3 and makes a W3. Therefore, PY has three possible moves (the defence of the W3) instead of one like before. But in any case the restricted game ends up in a PX victory, for example the one reported on the right.

\linespread{0.6}   
\begin{verbatim}
    1
++XX++                          +OXXX+                     
    X                               X                           
    X                               X                        
    +                               O                        
 ++X+X++                         ++XXX++                                       
    2
\end{verbatim}
\linespread{1.1}

\vspace{6pt}
\noindent
{\bf Comments.} As a first comment, note that any complex attack or victory can be analysed and constructed using either of the procedures that we have presented. In fact, any attack obtained applying the Lemmas can be regarded as a sequence of threats. And any sequence of threats can be decomposed and analysed using the Lemmas. Thus, the two procedures are just different points of view on the complex pattern construction.

As a second comment, note that a victory in the restricted game does not automatically translates into a victory in the full game. Concerning this point, the more the steps of a pattern, the higher the likely that the opponent can block the pattern in the full game. Therefore longer patterns are less useful. Moreover, a threat sequence composed by T2 (i.e. S3) only, like the first threat sequence example, is stronger than a sequence also containig T3, like the second threat sequence example. Indeed, in the first case PY only has one move per turn while in the second case it has more moves.

Finally, note that we cannot classify all the complex patterns, because they are too many. The best option to detect complex patterns is to use a tree search procedure: this approach will be discussed in section \ref{sec.player}.

\subsection{Board and moves analysis}
\label{vic.ana}

In this section we present a board analysis procedure that detects if a board is a sure win or a sure defeat. This is the minimal evaluation procedure that could be implemented in an artificial player. Moreover, the procedure also finds the possible player's moves. As a preliminary remark, note that in the discussion we may neglect the victories and consider only attacks and threats. In fact, if a V0 is on the board, the game is over and no analysis is needed. On the other hand, if a Vn with $n \geq 1$ is on the board, based on the Victory Decomposition Lemma, it can always be decomposed into two or more attacks.  

A general line of reasoning is as follows. Consider a PX board. If the board contains no attacks, no immediate victory is in sight. On the contrary, if the board contains PX and PY attacks, denote by $n_x$ and $n_y$ the number of steps of the quickest PX and PY attacks, respectively, setting it to $\infty$ if the player has no attacks. If $n_X \leq n_Y$, then, in principle, PX can play his attack and achieve a victory. On the contrary, if $n_X > n_Y$, PY has a faster path to the victory. In this situation PX has a limited set of moves: he must block the PY attack, which can be done either by playing in the attack defence or by placing on the board a faster PX attack. 

The problem with the latter general reasoning is that a victory in the restricted game does not always translates into a victory in the full game. Therefore, even if PX has an attack faster than PY, it cannot be guaranteed that his attack will really produce a victory: this depends of the full board status. As result an exact analysis for the general case is probably impossible. On the other hand when the number of steps of the attacks is suitably limited, an exact analysis becomes possbile. In the following we will present such an analysis, including attacks up to two steps. As a starting point, we give the following Lemma:

\begin{lemma} \label{lemma.min.dec} {\em {\bf One and two steps attacks decomposition.} Any one step attack (A1) contains at least an S4. Any two steps attack (A2) contains at least a W3.

\noindent
{\bf Proof:} The first statement is obvious. In order to prove the second statement, consider an A2. It cannot contain any S4 or S5, otherwise it would be an A1 or stronger. In the A2 it must exist an empty square such that, when filled with a PX stone, at least two S4 are produced, with a disjoint empty square (that is a D4), so that PX can surely win at his next move. Therefore, before placing the PX stone, there are two S3 intersecting in a single empty square. That is, a W3. } $\blacklozenge$ \end{lemma}

The last Lemma shows that by tracking all the S4 and the W3, we are sure to track all the A1 and the A2. In turn, tracking all the S4 and the W3, only requires to track all the B4 and the B3. 

We now proceed to analyse a PX board. We consider several mutually exclusive cases. The analysis determines if the board is a sure win or a sure defeat for PX and computes the possible moves for PX.

\vspace{6pt}

\noindent
{\bf 1) One or more S4X. } When there are one or more S4X on the board, the analysis is simple: PX plays and wins. 

\noindent
{\bf 2) One S4Y. } When there is one S4Y (and no S4X) on the board, again the analysis is simple: PX has to play in the defence of the S4Y, otherwise he will loose the game. Therefore PX has a single move. 

\noindent
{\bf 3) Two or more S4Y. } When there are two or more S4Y (and no S4X) on the board, PX has lost, because it cannot stop all the PY attacks\footnote{We are implicitely assuming that the S4Y have triggers in different squares. But this is true, otherwise the board would contain an irrational, impossible pattern.} neither he can play a faster attack.

\noindent
{\bf 4) One or more W3X. } When there are one or more W3X (and no S4X or S4Y) PX has won. In fact, there are no S4Y on the board, meaning that PY cannot win in his next move. Moreover, by playing any of the W3, PX will place two S4X on the board and PY will not be able to stop both.

\noindent
{\bf 5) One or more W3Y. } When there are one or more W3Y (and no S4X, S4Y, W3X) PX has a limited set of moves, discussed below. 

A first set of possible moves is constituted by triggers of the S3X. In fact, if PX plays an S3X it  will place an S4X on the board and PY will have to stop it and cannot play his attacks, which are two steps away from the victory. In this way PX can delay the victory of PY and, if he is lucky, by playing the S3 he can also disrupt the PY attack. An important exception is when, by stopping the S4X, PY can also play one of his W3Y attack: in this case the S3X would be useless and cannot be played. 

A second set of possible moves is given by the defences of the W3Ys. Since there can be more than one W3Y, by the Intersection Attack Lemma, we have to take the intersection of the defences of all the attacks. If the intersection is void, the attack is actually a victory and there is no defence.  On the other hand if the intersection is not void, it can be added to the possible set of moves. Note that, in this phase, a D3Y, having a defence of two squares only, would be detected.

\newpage
\section{Artificial players}
\label{sec.player}

In this section, we briefly discuss the main aspects that a developer must consider when writing an artificial Gomoku player. There are a number of different design choices, but the following features are common to any program. 

As a first building block, the artificial player relies on a data structure able to represent the {\bf board status}. This structure is not a simple matrix keeping track of the stones placement: it is tipically a much more complex data structure that also tracks patterns and ancillary data, allowing to implement the rest of the player's procedures in an efficient way. A second component of the player is a {\bf moves generation procedure}. Given a board, this procedure will generate all or a subset of the possible moves. A third component of the player is a {\bf board evaluation procedure}. Given a board, this procedure will rate it and produce a value that represents how good the board is for the player. This procedure must be able to recognise a vitory or a defeat, which are ranked as the best and worst board, respectively.

A basic player can be implemented using these three building blocks only and works as follows. Given the current board, the possible moves are generated. Then, each of the moves is tested. Specifically, the board after the move is produced and evaluated. The move producing the best board is selected as the actual move.

The latter approach would be satisfactory if the board evaluation procedure were ideal. In particular, if, for a given board, we could evaluate the probability of winning the game starting from that board\footnote{Gomoku is a deterministic game. Therefore, assuming rational players, for any given board the probability would be a binary value: one (victory) or zero (defeat or draw). If one or both players followed a partially random strategy, the probability would not be binary.}, the latter procedure would choose the move maximizing the winning probability and would realise the best possible player. 

Unfortunately, practical evaluation procedures are not ideal. Therefore, in order to improve the player, one has to develop a deeper exploration of the moves tree. That is, when evaluating a given move, we do not directly evaluate the resulting board. Instead, we generate all the counter moves of the opponent and produce the corresponding boards. For each of these boards we can further examine our moves and produce the corresponding boards. And so on up to a given depth. Only when the maximum depth is reached, the boards are evaluated and the best move is found. The latter approach can be implemented in many different ways and identifies a fourth key block of an artificial player, which is the {\bf tree search procedure}.

Finally, note that the artificial player is subject to some {\bf time and memory constraints}. Given these constraints, a trade-off between the tree search and the evaluation and generation procedures exists. Indeed, if the procedures are computationaly simple, a deeper search can be performed. However, if the procedures are too simple, their results could be inaccurate and the search could be less effective. As a result, the complexity and the quality of the evaluation and generation procedures is a key factor that should be carefully balanced when developing an artificial player.

\subsection{Minimax and Alpha Beta tree search}

\begin{figure}[t] 
  \begin{center} 
     \includegraphics[width=0.8 \linewidth, height = 0.25 \linewidth]{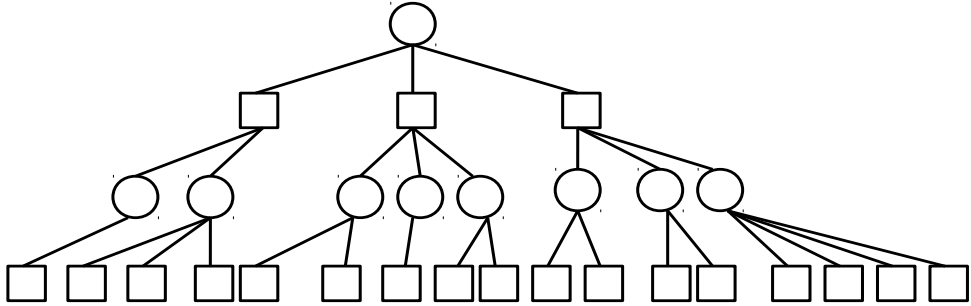} 
    \caption{\small The moves' tree. The root is the current board. The circles are maximising nodes, corresponding to PX boards. Their value is the Max of the values of their childs. The squares are minimising nodes, corresponding to PY boards. Their value is the Min of the values of their childs. If a node is a leaf, its value is obtained by running the board evaluation procedure. The edges represent the possible moves, produced by the moves generation procedure. } 
    \label{fig.tree} 
  \end{center} 
\end{figure}

The classical approach to the development of an artificial player is based on the Minimax tree search \cite{knuth}. In this approach, the tree is explored up to a given depth, as illustrated in figure \ref{fig.tree}. The root node represents the current board, belonging to PX. Each edge stemming from the root corresponds to one of the possible PX's moves and brings to a PY board/node. From each PY node, all the PY's moves can be explored and yield to a deeper level PX's node, and so on. 

In order to evaluate the possible moves we proceed as follows: we have an evaluation procedure that gives a value to the board from the point of view of PX, i.e. the higher the value the better the board for PX. This procedure is invoked when the last level (the deepest one) is reached. The board value is propagated from the deepest nodes (the leafs) to the root. In particular, assuming rational players, each PX node will try to maximise the board value and will therefore select the child having the highest possible value. In this sense, a PX node is said a Maximising node. On the contrary a PY node will try to minimise the board value and will therefore select the child having the least value. In this sense, a PY node is said a Minimising node and the whole procedure a Minimax search. Using this approach we can assign a value to each of the root moves and select the best one.  

\vspace{6pt} 
\noindent
{\bf Alpha Beta pruning and Principal Variation Search. } In order to increase the exploration depth, a key step is to develop a pruning method that prevents the search to waste time looking into bad moves and their subtree. The most famous pruning method is Alpha-Beta (AB) which is excelently described in the seminal paper from Knuth \cite{knuth}. The basic AB method can reduce complexity of the search by a factor two or higher. However, using more sophisticated approaches, one can further reduce the complexity. A remarkable improvement is the Principal Variation Search (PVS) exploiting the Iterative Deepening \cite{pvs}. In the latter method, the tree is generated one level per time and the best move sequence found at a given level (the principal variation) is explored first at the next level. Often the principal variation provides the best move also for the deeper level and in this case the move can be confirmed without fully exploring the alternatives.

\vspace{6pt} 
\noindent
{\bf Transposition Tables. }  Another way to increase the efficency is to note that different nodes of the tree can represent the same board. For example, starting from the root, assume that the following sequence of moves is performed: PX plays in $i$, PY plays in $j$ and PX plays in $k$.  Now consider the following, alternative sequence: PX plays in $k$, PY plays in $j$ and PX plays in $i$. It is clear that the two sequences produce to the same board. However, on the tree, the two sequences bring to different nodes, with different sub-trees. In this case, if we already explored one of the subtrees, it is useless to explore the other subtree, because it is identical to the first one. This waste can be avoided by the use of a Transposition Table (TT), which is a large database storing the boards already evaluated. Tipically, a TT is realised as an hash table, where the hash key is computed from the board \cite{pvs}.

\subsection{Monte Carlo tree search and deep learning}

As we have seen, in an artificial player a trade off exists between the complexity of the board evaluation and the depth of the search. Typically, in the Minimax search one tries to keep the board evaluation and the move generation procedures simple, in order to allow a deep search. Recently, the other side of the trade off gained momentum, thanks to the success of an artificial player for the game of Go, which is another board game, sharing some similarities with Gomoku. That artificial player, named Alphago \cite{alphago0,alphago}, was the first program capable of beating a professional human player. In Alphago the board evaluation is highly complex and is performed by means of a deep Convolutional Neural Network (CNN), trained on a large number of games by means of reinforcement learning.

The computational complexity of the CNN board evaluation is much higher than the complexity of a simple evaluation like those used in the Minimax approach and prevents a deep tree search. Indeed, Alphago does not perform a full tree search, up to a given depth. It simply explores some of the tree branches. The branches are selected randomly, with a probability that is modulated by the board value. Such a search is termed a Monte-Carlo tree search (MTS). 

The latter approach has been exported to the Gomoku game too. Several players were proposed which are an adaptation of Alphago to the Gomoku game, e.g. \cite{alphagomoku,alphagomoku5,alphagomoku4,alphagomoku3,alphagomoku2,alphagomoku0}. The best one is the Alphagomoku player \cite{alphagomoku}, which took part in the Gomocup tournament and ranked among the top ten players. However, Alphagomoku is still below the performance of other players based on the AB search.

\subsection{Endgame solvers}

A tool that may be useful in an artificial player is an endgame solver. Such a tool is used when a board that could be a sure victory is found. Clearly, such a board cannot be at the beginning of the game, instead it will appear at the end of the game. When such a board is found, the endgame solver shall explore the moves tree, trying to find a path towards the victory. The tool can be seen as a binary evaluation procedure of the board, stating if it is a sure victory or not. In the following we review three  endgame solvers. 

\vspace{6pt} 
\noindent
{\bf Binary Minimax (BMM).}  This solver explores the tree below a board containing one or more PX threats. It is based on a board evaluation procedure that classifies a PX board into three classes and specifically as a victory, a bad or a good board. The board is a victory if it is a sure victory for PX, for example if it contains an A1X. The board is good when no sure victory for PX can be identified but there are some threats on the board so that PX can play the threats and reduce the numbers of moves of PY. The board is a bad board if there are no sure victories neither threats. 

Based on the latter evaluation procedure, starting from a good PX board, we can explore the tree below the root using the Minimax search. When a victory or a bad board is found, the search is stopped and the value propagated towards the root. When a good board is found, the search is continued, but only considering the boards that can be obtained by playing the PX threats. In this way, eventually, the root board can be classified as either a victory or a bad board. This approach will be called a Binay Minimax (BMM) solver.

As a comment, note that the BMM performs a full search of the tree moves, which may seem unfeasible. However, the branching at each node is limited. In fact, in the PX nodes (Max), we only play the PX threats, and in the PY nodes (Min), the number of moves is limited because at least one PX attack is present on the board. Moreover, we can exploit pruning to reduce the search complexity: for example, if we find that a child of a Max node is a victory, we do not need to explore the other childs. Similarly, when we find a child of a Min node that is bad, we do not need to explore the other childs. Finally, we can exploit the TT to further reduce the search complexity. As a result, the full search performed by the BMM is usually feasible.

\vspace{6pt}  
\noindent
{\bf Proof number search (PNS).} Proof Number Search \cite{allis02,herik} is similar to BMM, in that it explores the subtree stemming from a board and aims at classifying the root board as either a victory or a defeat. Again, the board evaluation procedure provides a ternary result, that is: prooved (victory), disprooved (defeat) and open. When an open board is found, its child are explored. The evaluations are propagated up to to the root using the Minimax procedure. 

Note that PNS performs a full search of the tree, which can become unmanageable. In order to reduce the search complexity, pruning and TT can be used. Moreover, as its specific feature, PNS tries to reduce the search complexity by maintaining two numbers for each node, the proof and disproof number. These numbers are used to select the next branch to be explored in such a way as to speed up the search, see \cite{allis02,herik} for the details. 

In our experience, even using pruning, TT and proof numbers, if all the possible moves are considered, the PNS search complexity becomes unmanageable on a generic board. In practice the method can be used only near a victory or a defeat. As an alternative, one can use the classification already exploited for the BMM, that is, declare that a board is a defeat if there are no PX threats. In this way the PNS becomes similar to BMM. An important difference is that the proof numbers are still exploited in order to speed up the search. However, we verified that the impact of the proof numbers is limited and that the alternative version of the PNS and the BMM have basically the same performance.

\vspace{6pt}
\noindent
{\bf Threat space search (TSS). } Threat space search \cite{allis01} is again an endgame solver where the tree rooted in a board containing one or more PX threats is explored. As a key simplification, in the tree exploration, the PY player is allowed to place more than one stone on the board. In particular, when PX plays a threat, thereby placing an attack on the PY board, PY is allowed to place a stone in all the defence squares, even if there are more than one. This has the important effect of reducing the branching factor. Moreover, if a sure PX win can be obtained in this way, it will be there also in the real game, where PY can place only one stone per turn.

The TSS is started from a board where a set of one or more PX threats has been identified. Each of the threats is played by PX, giving rise to a set of PY boards where at least an attack is present. Then the defences of the attack are filled with PY stones. And if any new PX threat, not belonging to the initial set, has been produced, the exploration is pushed forward, by playing all the newly created threats, until a board with no new threats is reached. During the exploration, the threat triggers are tracked because if two threats with the same trigger are detected, based on the Combination Attack Lemma, these are potentially an attack. Even better, this is a potential victory for PX, because the attack is on a PX board. However, the attack needs to be confirmed: first by checking that the threats are compatible and next by checking that the defences are disjoint. Moreover one has to check that PY cannot achieve a faster victory when the moves yielding to the attack are played in the full game. See \cite{allis01} for the details. 

Thanks to the fact that PY is allowed to place more than one stone, the tree explored by the TSS has a lower branching factor than the full tree, making the search extremely efficient. However, the need to track the triggers and to confirm the attacks adds a great deal of complexity to the procedure. Overall, we found that the performance of TSS are on the order of that of BMM. As an additional drawback, since PY is given the advantage of placing more than one stone, some PX wins may be missed.

\vspace{6pt}
\noindent
{\bf Comments. } As a general comment note that for the endgame solvers that are based on the PX threats, like BMM and TSS, one can produce different solvers by varying the set of threats used. To clarify the latter point consider, for example, the TSS. The solver can exploit only the T2 patterns (i.e. the S3s) or it can exploit both the T2 and the T3 patterns (or a subset of the T2 and T3). In the first case, the code is simpler to develop and the search is faster to run. In the second case the code and the search are more complex, but more options are explored and a wider range of attacks can be found. This is again a trade off between complexity and performance. In principle, we could even consider a TSS which exploits more treats, for example T2, T3 and T4, but the code would become extremely complex and the computational complexity very high. In summary, when implementing an endgame solver, the choice of the set of threats that are used is an important choice, that has to be carefully evaluated.

\vspace{6pt}
\noindent
{\bf Using the solver. } To conclude, we briefly discuss how the endgame solver can be used in an artificial player and highlight some difficulties in its use. A first option is to use the endgame solver as a first step in the analysis, that is, to run the solver on every board that we encounter. A problem with this approach is that, if the board is not a win, the solver will just waste some computation time and we need to run the AB search after the solver in order to select a move. On the other hand, if the board is a sure win, usually both the solver and the AB search will detect the victory. Surely, the solver will be much faster than AB search, however this advantage comes only once per game, at the very last analysed board. As a result, running the solver on any board we encounter is a questionable strategy.

As an alternative, we could run an AB search up to a given depth and then run the solver from any leaf of the AB tree where one or more threats are present. This may be a better strategy but requires some caution too, because in order to make room for the solver time, we need to limit the depth of the AB search. Since there are many leafs that need to be analysed, the solver will run many times and the maximum AB depth can be severly limited by this approach, which is not necessarily a good choice. Moreover, even if we find a leaf that is a sure win, usually we cannot guarantee that we can reach that leaf starting from the root, so that the computing power dedicated to the solver may be not justified. Essentially, in this case we are using the solver as an evaluation function and we have to consider the complexity-depth trade off.

In practice, we are not aware of any top level Gomoku artificial player using an endgame solver.

\newpage
\section{Wine}
\label{sec.wine}

Wine is an artificial Gomoku player developed by Jinjie Wang. It is written in C++ language and the source code is available on Github \cite{wine}. Wine is a strong player: it ranked tenth in the 2020 edition of the free style tournament of the Gomocup. Moreover, the code is well written and the brain is comparatively simple. As a result, Wine is an excelent starting point for anyone wishing to understand the structure and organization of a modern Gomoku player, which is a prerequsite to develop a new player. Unfortunately, there is no documentation about the program, except for some coments in the code, mostly in Chinese language. This means that, in order to understand the Wine functioning, one must embark in a difficult reverse engineering task. In the last year we have accomplished this task and in this section we give a presentation of Wine, which can be of help to anyone willing to understand the code. Moreover, we push the analysis a bit further and present some possible improvements to the code that can increase its strength. In the dicussion we will focus on the data structures used for the board representation and on the move generation and board evaluation procedures. Concerning the tree search procedure, Wine employs a PVS search with Iterative Deepening and TT, which is excelenty discussed in \cite{pvs}. 

\subsection{Lines}

We introduce a special pattern, namely the {\bf line}. This is a pattern constituted by nine aligned squares (horizontally, vertically or diagonally) and is exploited by Wine and other players to organise and track the game patterns. 

If the central square of the line contains a PX stone, the line is said a {\bf PX line}. Note that a line contains five blocks and the central square of the line is contained in all these blocks. Thus, if the central square contains a PX stone, PY will never be able to build a victory in that line. Also note that a PX line is different from a PX pattern as defined earlier. For example the following line \verb|OOXXX+XO+| is a PX line but not a PX pattern and the following line \verb|++XX++XX+| is not a PX line but it is a PX pattern.

To proceed, we introduce some types of line, containing the patterns introduced in section \ref{pat.exe}. To simplify the naming, we will use the same name for the line and for the pattern that it contains: the context will make clear if the name refers to the pattern or to the line. The lines are presented below, in order of decreasing strength, from the strongest to the weakest. It is understood that when a line could be assigned to two or more different classes, it belongs to the strongest one\footnote{For example, a double four line also fulfils the definition for a simple four line. But it is classified as a double four.}.

\vspace{6pt}

\noindent
{\bf $\bullet$ Simple five.} A PX line containing one or more S5. Equivalently, a PX line containing a V0. E.g. \verb|+OXXXXXOO|, \verb|+OXXXXXX+|. 

\noindent
{\bf $\bullet$ Double four.} A PX line containing a D4. Equivalently, a PX line containing a V1. E.g. \verb|+O+XXXX+O|, \verb|XXX+X+XXX|.

\noindent
{\bf $\bullet$ Simple four.} A PX line containing an S4. Equivalently, a PX line containing an A1 with size one defence. E.g. \verb|+O+XXXXOO|, \verb|XXXOX+XXX|.

\noindent
{\bf $\bullet$ Double three.} A PX line containing a D3. Equivalently, a PX line containing an A2 with size two defence. E.g. \verb|+++XXX++O|, \verb|++XXX+++O|.

\noindent
{\bf $\bullet$ Weak three.} A PX line containing a W3. Equivalently, a PX line containing an A2 with size three defence. E.g. \verb|+O+XXX++O|, \verb|O+XXX+++O|.

\noindent
{\bf $\bullet$ Simple three.} A PX line containing an S3. Equivalently, a PX line containing a T2. E.g. \verb|+O+XXX+OO|, \verb|OOXXX+++O|.

\noindent
{\bf $\bullet$ Double two.} A PX line containing a T3 with a trigger that, when filled with a with a PX stone, turns the line into a double three. E.g. \verb|++++XX++O|, \verb|O++XX+++O|.

\noindent
{\bf $\bullet$ Weak two.} A PX line containing a T3 with a trigger that, when filled with a with a PX stone, turns the line into a weak three. E.g. \verb|+O++XX++O|, \verb|O++XX++OO|.

\noindent
{\bf $\bullet$ Simple two.} A PX line containing an empty square that, when filled with a with a PX stone, turns the line into a simple three. E.g. \verb|+O++XX+OO|, \verb|OO+XX++OO|.

\noindent
{\bf $\bullet$ Generic.} A line not belonging to one of the previous classes is termed a generic line.

\vspace{6pt}

For the sake of precision, we note that Wine uses a coarser classification. In fact, Wine makes no difference between a double/weak three and a double/weak two. It only considers two patterns (termed a flex three and flex two) to represent the four cases. However, the double and weak threes have different strenghts. In particular the double three is stronger because it has a smaller defence. Thus this finer classification can be used to improve the player.

\subsection{Line classification}

Having introduced several types of line, we need a classification procedure. Such a procedure can be found in the Wine code\footnote{The function {\em LineType} and its subfunctions.}, but is not adequate for our classes because, as we have said, we are considering a finer classification. Therfeore, in the following, we discuss an alternative.

To proceed, note that in a line there are five blocks. We denote by $G$ the maximum degree of the PX blocks and by $N$ the number of blocks of degree $G$. Next, given any two blocks of degree $G$, say $A$ and $B$, we consider a vector which will be called a tracker vector and is defined as follows. It is a nine elements vector where each element is orderly associated to one of the squares of the line. The value of each element counts the number of empty squares in the two blocks. In other words, the $n$-th element is zero, one or two: it is zero if the $n$-th square does not belong to the blocks or is filled in both blocks; it is one if the $n$-th square is an empty square in one of the two blocks; and is two if the $n$-th square is an empty square in both blocks. Note that, since there are $N$ blocks of degree $G$, there are $N(N-1)/2$ different trackers for the line. Below we make some examples:

\linespread{0.6}   
\begin{verbatim}
 =====      A                      =====        A
0+XXX++00   Line G=3 and N=2       ++XXX++++    Line  G=3 N=3
  =====     B                       =====       B
010002100   Tracker                120001000    Tracker                        
\end{verbatim}
\linespread{1.1}

In order to classify the line, we use $G$ and $N$ and consider their possible values. In this way the classification can be broken down into several mutually exclusive cases discussed below.

\noindent
{\bf Case 1) } If $G=5$, the line is a simple five.

\noindent
{\bf Case 2) } If $G=4$ and $N=1$, the line is a simple four. Its defence is the empty square of the S4.

\noindent
{\bf Case 3) } If $G=4$ and $N>1$, we need to inspect the tracker vectors. Note that, when $G=4$, the sum of the elements of the tracker is two and there are two types of vectors: 1) those with two elements equal to one and the others zero and 2) those with one element equal to two and the others zero. Below we make some examples:

\linespread{0.6}   
\begin{verbatim}
 =====      A                        =====     A
0+XXXX+00   Line G=4 N=2           OOXXX+XX0   Line G=4 N=2
  =====     B                         =====    B
010000100   Tracker                000002000   Tracker 
\end{verbatim}
\linespread{1.1} 

\noindent
When there is at least one tracker of type 1) there are two S4 with disjoint defences, i.e. a D4, and the line is classified as a double four. On the contrary if all the trackers are of type 2) the line contains no victories and is a simple four, the defence of which is given by the only non zero square of the tracker vectors. Below, we make two more examples:

\linespread{0.6}   
\begin{verbatim}
 =====      A                        =====      A
X+XXXX+XX   Double four            OXXX+XXXO    Simple four
  =====     B                         =====     B
010000100   Tracker                000020000    Tracker                              
\end{verbatim}
\linespread{1.1}

\noindent
{\bf Case 4) } If $G=3$ and $N=1$, the line is a simple three.

\noindent
{\bf Case 5) } If $G=3$ and $N>1$, we need to inspect the tracker vectors. Note that, when $G=3$, the sum of the elements of the tracker is four and there are three types of vectors: 1) those with two elements equal to one, one equal to two and the others zero; 2) those with four elements equal to one and the others zero; and 3) those  with two elements equal to two and the others zero. Below we make some examples:

\linespread{0.6}   
\begin{verbatim}
 =====      A           =====      A              =====   A
0++XXX+00   G=3         XX++X++XX  G=3         O+OXXX++X  G=3
  =====     B               =====  B               =====  B
012000100   Tracker     001101100  Tracker     000000220  Tracker                  
\end{verbatim}
\linespread{1.1} 

\noindent
Note that when the tracker is of type 1) the two blocks form a weak three, that is an A2. Indeed by playing where the vector is 2, PX can produce two S4 with separate defences and PY cannot stop both. When the vector is of type 2) PX cannot produce two B4 with a single move so that the pair is not an attack, only a threat, namely a simple three. When the vector is of type 3), PX can produce two S4 with a single move, by playing where the vector is 2, but PY will stop both of them by playing in the other non-zero square of the vector, so this is again a simple three.

In summary, if there are no vectors of type 1) the line is a simple three. If there is one vector of type 1) the line is a weak three, with a defence given by the non zero squares of the vector. When there are two or more vectors of type 1) the line is an attack with a defence given by the intersection of the non-zero elements of the vectors and can be a weak three (defence of size 3) or a double three (defence is size 2). Below, we make an example:

\linespread{0.6}   
\begin{verbatim}
0++XXX++X  G=3   N=4   (N-1)N/2=6
 =====     A
  =====    B
   =====   C
    =====  D
012000100  AB    A2
001000210  BC    A2
011000110  AC
011000110  AD
001000210  BD    A2
000000220  CD
--------------------
001000100  A2 defences intersection: line is D3
\end{verbatim}
\linespread{1.1} 

\noindent
{\bf Case 6) } If $G=2$ and $N=1$, the line is a simple two.

\noindent
{\bf Case 7) } If $G=2$ and $N>1$, we proceed as follows. If, by filling one of the empty squares, we are able to produce a double three, the line is a double two. Otherwise, if we are able to produce a weak three, the line is a weak two. Otherwise the line is a simple two.

\noindent
{\bf Case 8) } If $G<2$ the line is a generic line.

\vspace{6pt}

As an important observation, note that we do not really need to run the latter classification procedure during the match. To elaborate, note that the line has nine squares, which can take on three values, namely {\em black}, {\em white} and {\em void}. In practice, it may be convenient to consider lines that extend outside the board too. This can be done by adding a fourth value, namely {\em outside}. Since four values can be represented by two bits, a PX line (having the central square assigned to PX) can be represented by 16 bits. Therefore, there are $2^{16}$ different lines. These can be indexed by an integer ranging from 0 to $2^{16}-1$. Then, the classification procedure is run only once, at the code initialization, and all the $2^{16}$ possible lines are classified. The results are stored in a table with $2^{16}$ elements. When, during the match, a line of the board needs to be classified, we do not need to run the classification procedure: we simply need to produce the 16 bit number from the line contents and use that number as an index to access the table.

\subsection{Board representation and potential lines}

The board representation is a critical part of every artificial player. It shall be simple and easy to update when a new stone is placed on the board. In addition, it should allow the player to efficently carry out three basic tasks, namely victory/defeat detection, moves generation and board evaluation. 

Having introduced the lines, an obvious way to represent the board status is to track the lines present on the board. In particular, each square is at the center of four lines, in the four directions, so that the board can be represented by $ 4 N_r N_c $ lines\footnote{Some lines will partially extend outside the board. As we have mentioned, this can be handled using a special content, indicating an {\em outside} square.}. However, Wine follows an elegant, slightly different approach. Specifically, instead of considering {\em the lines already on the board}, Wine considers the {\em lines that can be produced by filling the empty squares of the board}. This approach is better described in the following.

To proceed, it is convenient to introduce the concept of {\bf potential line}, which is a line having an empty central square. Therefore a potential line can become both a PX or a PY line, depending on which stone is placed in its central square. Accordingly, a potential line can be evaluated for PX and for PY, by assuming that either PX or PY places a stone in the central square. For example, the following line \verb|+XXX+OOO+| can be used by both PX and PY to make a simple four. And the following line \verb|+XXX++OO+| is a potential double four for PX and a potential simple three for PY.

As we mentioned, in order to represent the board, Wine tracks the potential lines instead of the actual lines. To some extent, this makes no real difference, since we could attach both a potential and an actual value to each line and the two are related. For example, the following line \verb|+XXX+OOO+|, which is a potential simple four for both PX and PY, could equivalently be considered an actual simple three for both players. The key point is that using the potential lines we shift our attention from the actual lines, which are four per square, to the potential lines, which are four per {\em empty} square. Therefore, the potential lines are less than the actual lines. As a result, they are simpler to track. Moreover, the potential lines are directly linked with the possible moves, which are the empty squares.

In practice, for each empty square, Wine tracks the potential lines that can be produced in the four directions by filling the square with a PX or with a PY stone. Therefore it maintains eight potential lines for each empty square. For example the central square of following line \verb|XXXX+OOO+| is a simple five for PX and a simple four for PY. The square has six more types on the other three directions. These potential lines are updated as soon as a move is made. As we see in the next subsections, these data are used to evaluate the board and generate the moves. 

Note that Wine also keeps other data regarding the board. Obviously, Wine tracks the content of each square (black, white or empty). Moreover, for each square, Wine keeps a flag indicating if the square is considered a possible move or is excluded a priori from the possible moves. In order to be a possible move, a square must be empty and must be close to an already filled square (maximum two squares apart). The latter condition is sensible, because playing in an isolated square is useless, and is enforced by most players. The condition narrows the set of possible moves thereby reducing the search complexity.  

Since Wine relies on the potential lines, from now on, if not stated otherwise, when we speak of a line, we mean a potential line.

\subsection{Cross patterns}

In addition to the lines we can also consider some useful cross patterns. In particular, in section \ref{pat.exe} we considered three types of cross patterns, namely the C44, C43 and C33. The corresponding potential patterns are obtained when a square is the center of two appropriate potential lines. In particular, when a square is crossed by two S4 lines, it is a (potential) {\bf cross C44}. Similarly, when a square is crossed by an S4 and a D3 or W3 lines, we obtain a (potential) {\bf cross C43}. Finally when a square is crossed by two W3 or D3 lines, it is a (potential) {\bf cross C33}. Again we may specify the nature of the three (weak or double) if needed. Below we report some examples.

\linespread{0.6}   
\begin{verbatim}
   C44                   C43w                 C3w3d

    +                      +                   +         
    +                      +                   +         
    +                      +                   O         
    O                      O                   +        
+XXX+O+++              +XX+++O++           ++XX+++O+                        
    X                      X                   +         
    X                      X                   X         
    X                      X                   X         
    +                      +                   +         
\end{verbatim}
\linespread{1.1} 

Clearly, these are important patterns. Indeed, by filling the empty square, PX places a victory on the board. In particular, a one step victory in the C44 case and a two steps victory in the C43 and C33 cases. 

We note that cross patterns involving three or four lines could be considered too. However these are irrational patterns and are unlikely to appear in a real game. The same is true for cross patterns made by two or more S5. Therefore, we do not take these patterns into consideration. In any case, if in a game there are three or more potential attacks in the same square, we shall consider the strongest pair. For example, if an empty square is an horizontal S4, a vertical W3, a main diagonal S4 and a secondary diagonal S2, it will be classified as a C44.

In summary, for each empty square and for each player, we shall track four potential line patterns, on the four directions, plus a single, potential cross pattern. We also note that Wine does not track all the cross patterns: it tracks the C44 but not the C43 and the C33. This does not mean that Wine is unable to discover these victories. In fact, Wine will discover these victories by means of the tree search. However, a direct tracking is more efficient.

\subsection{Defence of some potential pattern}

In the next section we will adapt the board analysis carried out in section \ref{vic.ana} to the potential patterns. To this end it is useful to discuss the defence for some of the potential lines. 

To start with, consider a potential S5 line, e.g. \verb|+XXX+X+++|. It contains an actual S4 pattern, which is an A1. If PY find this line on his board, he can stop the PX attack by playing into the line's central square only. Therefore, an S5 has a defence of size one, constituted by the central, empty square.

Consider now a potential D4 line: it contains an actual W3, which is an A2 with a size three defence. The D4 can be stopped by playing in the defence of the W3. Therefore, a potential D4 has a size three defence. Below we make two examples. 

\linespread{0.6}   
\begin{verbatim}
++XX+X+++   Potential D4 containing an actual W3
 -  - -     Its defence

O+XX+X+O+   Potential D4 containing an actual W3
 -  - -     Its defence
\end{verbatim}
\linespread{1.1} 

It is instructive to study how an actual D3, which is an A2 with a size two defence, is recognised using the potential lines. When a D3 is on the board, it will be contained in two potential D4. Therefore, two attacks (the potential D4) with defence size three will be recognised. Below we make an example:

\linespread{0.6}   
\begin{verbatim}
  ===-*===-            Potential D4 line (trigger and defence marked)
 +++O++XXX++O+++       Actual D3 pattern
      -===*-===        Potential D4 line (trigger and defence marked)
\end{verbatim}
\linespread{1.1} 

\noindent
In the latter case, by using the Intersection Attack Lemma we take the intersection of the defences of the two D4 and find that the union of the two has a defence of two squares, coincident with that of the actual D3.

Finally, consider a potential C44: it contains an actual cross W3, which is an A2 with a size three defence. The D4 can be stopped by playing in the defence of the cross W3. Therefore, a potential C44 has a size three defence. Below we make an example:

\linespread{0.6}   
\begin{verbatim}
   C44       
    +         potential C44
    +         contains actual cross W3
    +       
    O       
+XXX+O+++     defence is the intersection square
-   X         plus the squares near the dashes
    X       
    X       
    +-       
\end{verbatim}
\linespread{1.1}

\subsection{Board and moves analysis}
\label{vic.ana2}

In this section we present an analyis of the victories and of the possible moves based on the potential lines which is obtained from the one reported in section \ref{vic.ana}. In particular, we obtain the new analysis by replacing the actual patterns S4, W3 and S3 of the previous analysis with a potential S5, D4 and S4 respectively. In the following, we analyse a PX board and consider several mutually exclusive cases.

\vspace{6pt}
\noindent
{\bf 1) One or more S5X. } When there are one or more S5X on the board, the anaysis is simple. PX has won and can play anyone of the S5X. 

\noindent
{\bf 2) One S5Y. } When there is one S5Y (and no S5X) on the board, again the analysis is simple. PX has to play in the S5X defence, i.e. the square where the S5Y is, otherwise he has lost. Theferore PX has a single move. 

\noindent
{\bf 3) Two or more S5Y. } When there are two or more S5Y (and no S5X) on the board, PX has lost, because it cannot stop all the PY attacks\footnote{We are implicitely assuming that the S5Y are in different squares. But this is always true, otherwise the board would contain an irrational pattern. Indeed, suppose, for example, that there are two S5Y. These cannot be on the same square. In fact, if they were on the same square, on different lines, when PY made that last S5Y he would have already had the other S5Y and therefore could have win earlier: this is not possible because PY is a rational player.} neither he can play a faster attack.

\noindent
{\bf 4) One or more D4X or C44X. } When there are one or more D4X or C44X (and no S5X or S5Y) PX has won. In fact, there are no S5Y on the board, meaning that PY cannot win in his next move. Moreover, by playing any of the D4 or the C44, PX can place two S5X on the board and PY will not be able to stop both.

\noindent
{\bf 5) One or more D4Y or C44Y. } When there are one or more D4Y or C44Y (and no S5X, S5Y, D4X and C44X) PX has a limited set of moves. 

A first set of possible moves, is constituted by the S4X. In fact, the potential D4Y and C44Y are two steps victories, and if PX plays an S4X he  will place an S5X on the board and PY will have to stop it and cannot play his attacks. In this way PX can delay the victory of PY and, if he is lucky, by playing the S4 it can also disrupt the PY attack. An important exception is when, by stopping the S4X, PY can also play one of his D4Y or C44Y attack: in this case the S4X cannot be played! In practice we can consider as possible moves all the S4X and let the tree search identify those which cannot be really played.

A second set of possible moves is given by the defences of the PY attacks. Since there can be more than one D4 or C44, by the Attack Intersection Lemma, we have to take the intersection of the defences of all the attacks. If the intersection is void, the attack is actually a victory and there is no defence. If the intersection is not void, it can be added to the possible set of moves.

\vspace{6pt}
\noindent
{\bf Comment.} The board analysis just presented is better than the one exploted by Wine. Indeed, Wine only considers a subset of the cases that we considered and relies on the tree search to spot the others, which is less efficient. 

\subsection{Board evaluation and move generation}
\label{sec_bd_eval}

When the tree search reaches a board which is not a victory or a defeat, two cases are possible. Either the maximum depth has been reached or not. In the first case the board needs to be evaluated. In the second case, a set of possible moves has to be generated, in order to push the tree search deeper. Therefore we need both a board evaluation and a move generation procedure.

These procedures are realised in Wine and are described in the following from the PX's point of view. The procedures are based on the board data kept by Wine, namely the potential lines of both players for each empty square and the set of possible moves. Moreover, Wine scores each line type according to its strenght, giving higher scores to the stronger lines. These scores are stored in a vector and are exploited by the two procedures.

The board value is obtained in a very simple way, namely by scanning the possible moves and by summing up the scores of their lines, giving a positive weigth to the PXs lines and a negative weight to the PY ones. Note that Wine does not account for the C43 and C33 patterns. Taking into account and scoring these patterns slightly improves the player.

Concerning the move generation procedure, it is very simple too. Again each line is assigned a score, not necessarily the same one used for the board evaluation. Next, the possible moves are scanned: for each possible move the sum of the scores of the lines that can be realised by playing the move is computed. In this case both the PX and the PY lines are given a positive weigth, because it is equally important to play a square if it is a good move for PX or if it kills a good move for PY. Finally the possible moves are sorted according to the their scores and the best $B$ are selected as possible moves. The parameter $B$ determines the branching factor of the search and is an important one. If the branching factor is low the tree search will go deeper. However if $B$ is too low some good moves could be not considered and are lost. The default value for Wine is $B=40$.



\newpage

\end{document}